\documentclass[10pt,twocolumn,letterpaper]{article} 

\usepackage{avss}
\usepackage{times}
\usepackage{epsfig}
\usepackage{graphicx}
\usepackage{amsmath}
\usepackage{amssymb}
\usepackage{wrapfig}
\usepackage{placeins}
\usepackage{tabularx} 
\usepackage{fancyhdr}

\avssfinalcopy 

\def\avssPaperID{77} 

\ifavssfinal\fi
\begin{document}

\renewcommand{\headrulewidth}{0pt}

\newcommand{\fig}[1]{Figure~\ref{fig:#1}}
\newcommand{\sect}[1]{Section~\ref{sect:#1}}
\newcommand{\tab}[1]{Table~\ref{tab:#1}}
\newcommand{\alg}[1]{Algorithm~\ref{alg:#1}}
\newcommand{\eq}[1]{(\ref{eq:#1})}

\newcommand{\p}{\mathcal{P}}
\newcommand{\e}{e}
\newcommand{\E}{E}
\newcommand\cincludegraphics[2][]{\raisebox{-0.45\height}{\includegraphics[#1]{#2}}}

\newcommand{\Sp}{{\mathcal S}^{+}}
\newcommand{\Sgen}{{\mathcal S}}
\newcommand{\Sm}{{\mathcal S}^{-}}
\newcommand{\spr}[2]{{\langle{}#1,#2\rangle}}
\newcommand{\prp}{p^{+}}
\newcommand{\prm}{p^{-}}
\newcommand{\Hp}{H^{+}}
\newcommand{\Hm}{H^{-}}
\newcommand{\hp}{h^{+}}
\newcommand{\hm}{h^{-}}

\title{Multi-region Bilinear Convolutional Neural Networks for Person Re-Identification}

\author{Evgeniya Ustinova\\
Skolkovo Institute of Science and Technology\\
Skolkovo, Moscow\\
{\tt\small evgeniya.ustinova@skolkovotech.ru}
\and
Yaroslav Ganin\\
Skolkovo Institute of Science and Technology \\
Skolkovo, Moscow\\
Montreal Institute for Learning Algorithms \\
Montreal, Quebec\\
{\tt\small yaroslav.ganin@gmail.com}
\and
Victor Lempitsky\\
Skolkovo Institute of Science and Technology\\
Skolkovo, Moscow\\
{\tt\small lempitsky@skoltech.ru}
}

\maketitle

\thispagestyle{fancy}


\fancyhf{} 


\begin{abstract}
In this work we propose a new architecture for person re-identification. As the task of re-identification is inherently associated with embedding learning and non-rigid appearance description, our architecture is based on the deep bilinear convolutional network (Bilinear-CNN) that has been proposed recently for fine-grained classification of highly non-rigid objects. While the last stages of the original Bilinear-CNN architecture completely removes the geometric information from consideration by performing orderless pooling, we observe that a better embedding can be learned by performing bilinear pooling in a more local way, where each pooling is confined to a predefined region. Our architecture thus represents a compromise between traditional convolutional networks and bilinear CNNs and strikes a balance between rigid matching and completely ignoring spatial information. We perform the experimental validation of the new architecture on the three popular benchmark datasets (Market-1501, CUHK01, CUHK03), comparing it to baselines that include Bilinear-CNN as well as prior art. The new architecture outperforms the baseline on all three datasets, while performing better than state-of-the-art on two out of three. The code and the pretrained models of the approach can be found at \url{https://github.com/madkn/MultiregionBilinearCNN-ReId}.
\end{abstract}

\section{Introduction}

\begin{figure*}[t]
\centering

\begin{tabular}{c c}

\begin{tabular}{c c c c}

\includegraphics[ height=2cm, width=1.2cm]{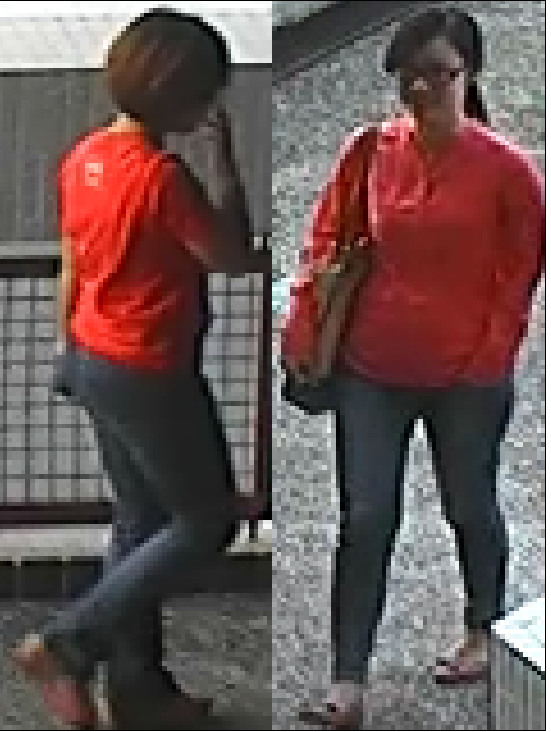}&
\includegraphics[ height=2cm, width=1.2cm]{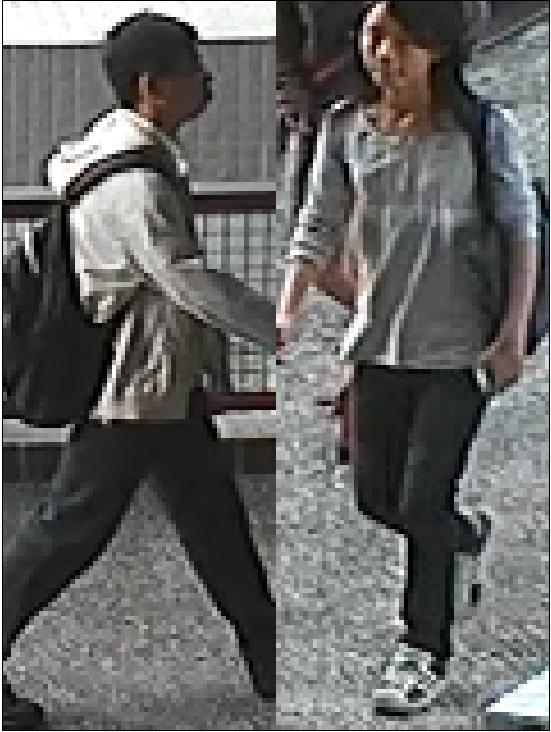}&
\includegraphics[ height=2cm, width=1.2cm]{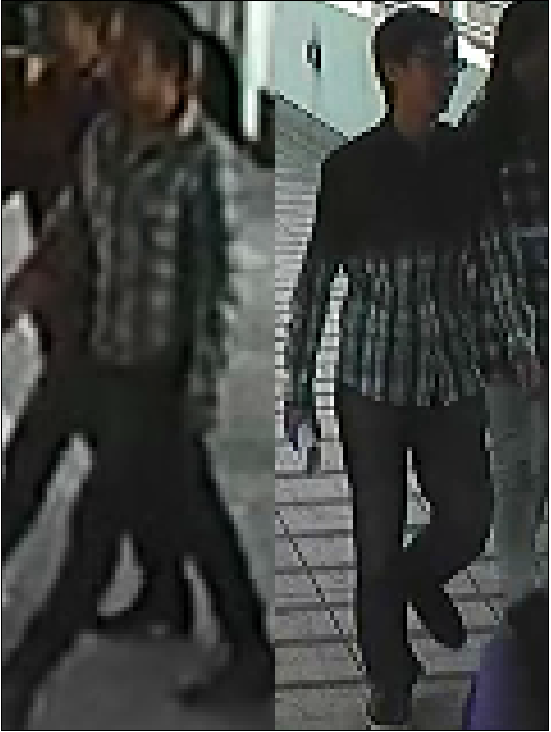}&
\includegraphics[ height=2cm, width=1.2cm]{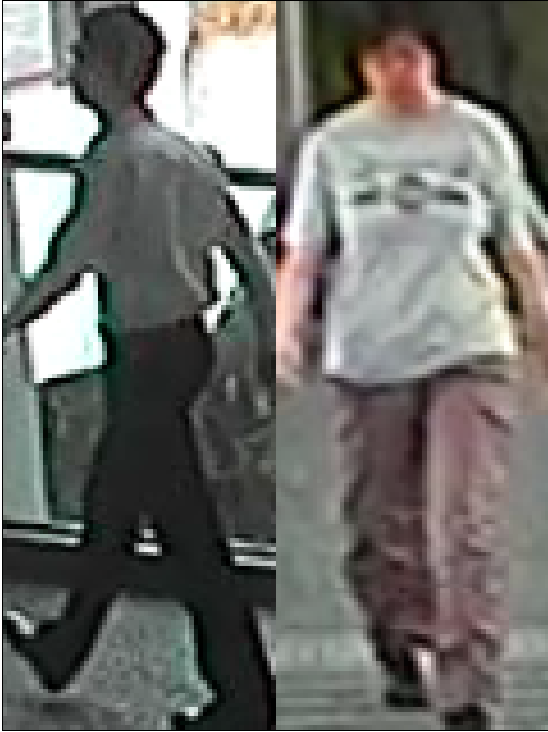}
\\
\includegraphics[ height=2cm, width=1.2cm]{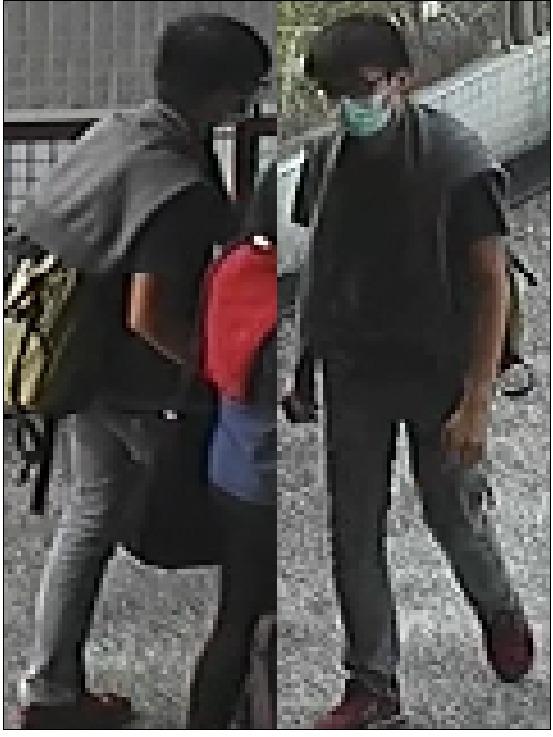}&
\includegraphics[ height=2cm, width=1.2cm]{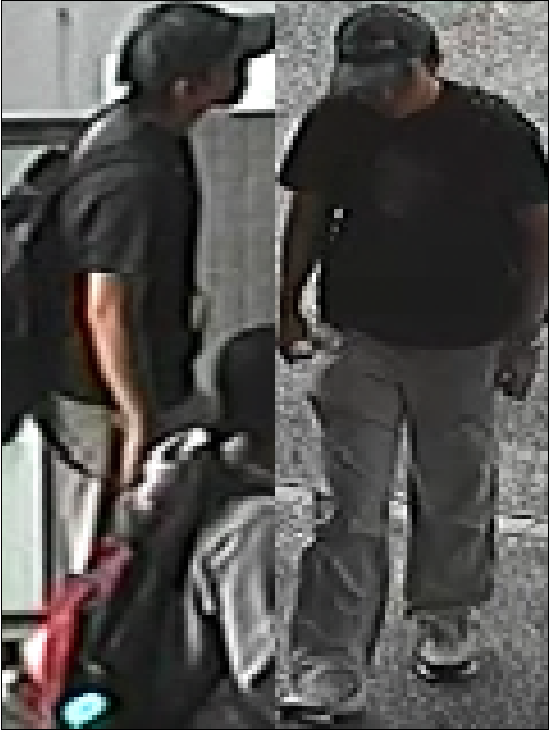}&
\includegraphics[ height=2cm, width=1.2cm]{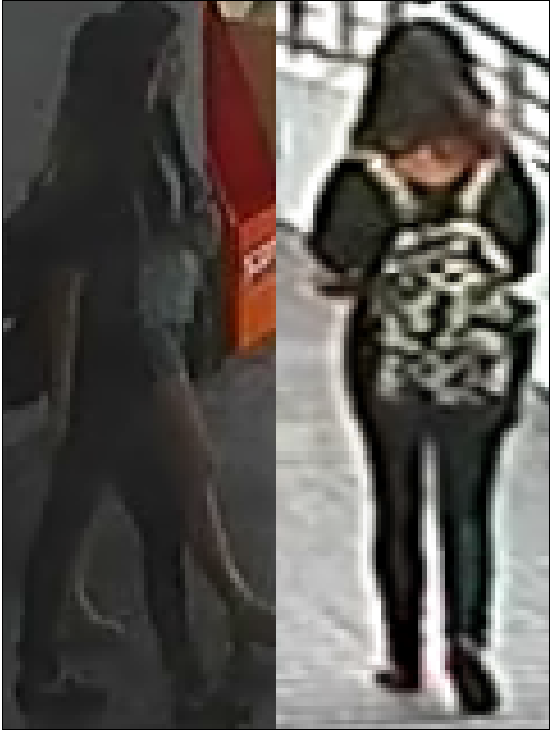}&
\includegraphics[ height=2cm, width=1.2cm]{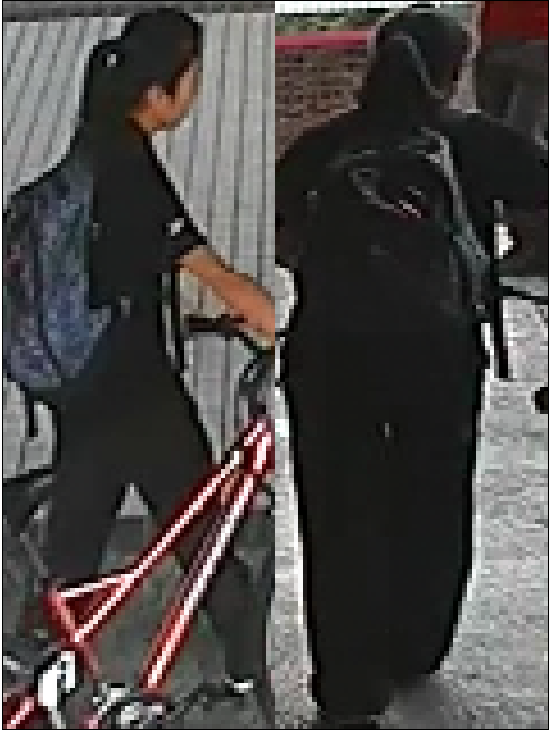}
\end{tabular} &
\begin{tabular}{c c}
\includegraphics[ height=2cm, width=4cm]{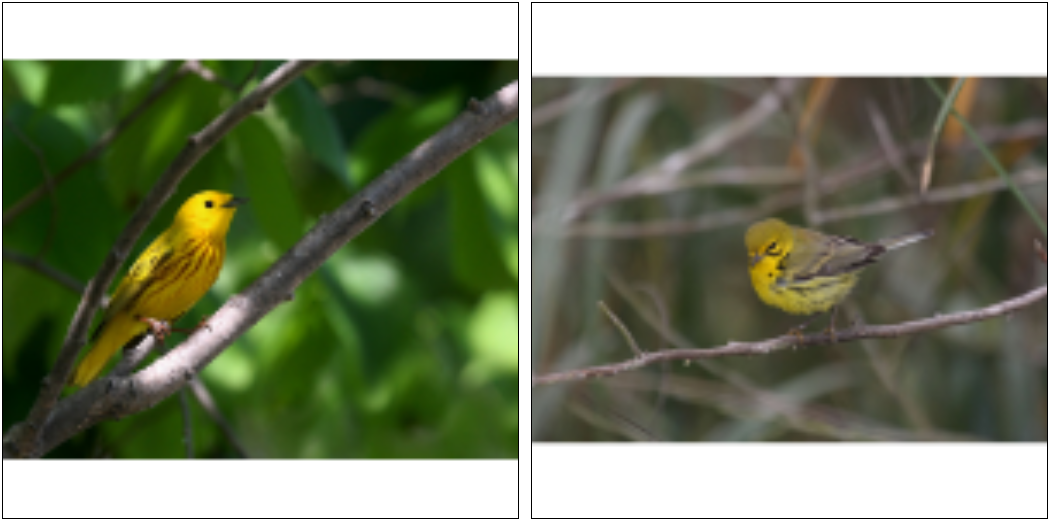}&
\includegraphics[ height=2cm, width=4cm]{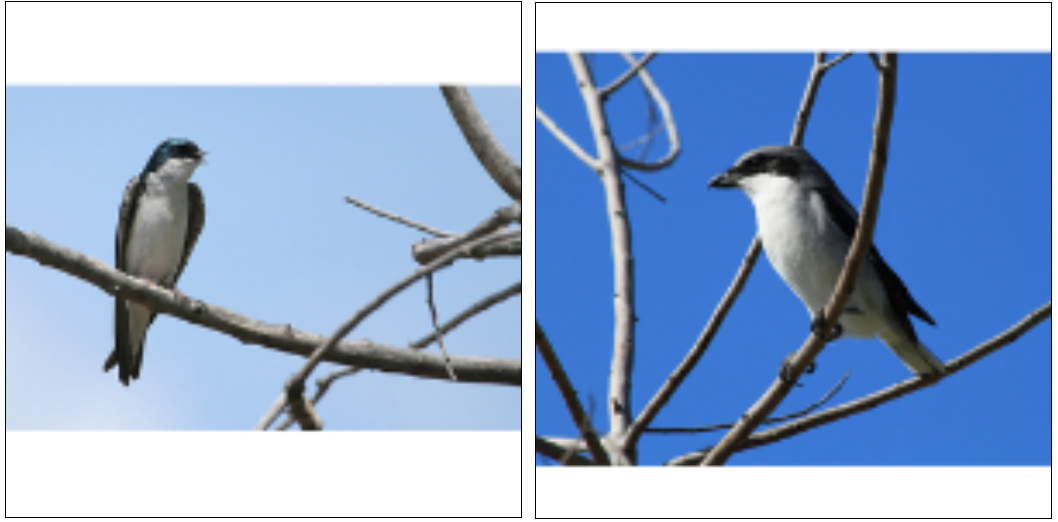}

\\
\includegraphics[ height=2cm, width=4cm]{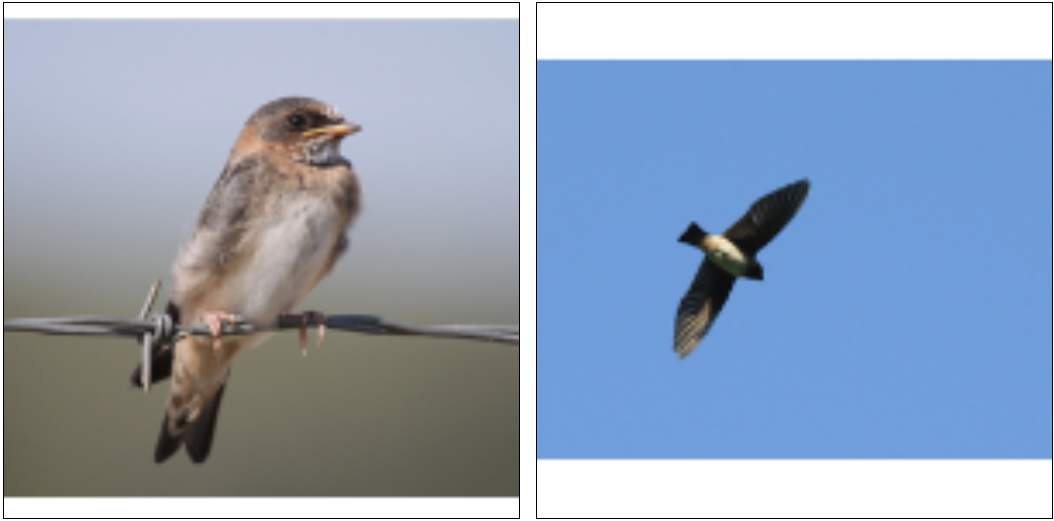}&
\includegraphics[ height=2cm, width=4cm]{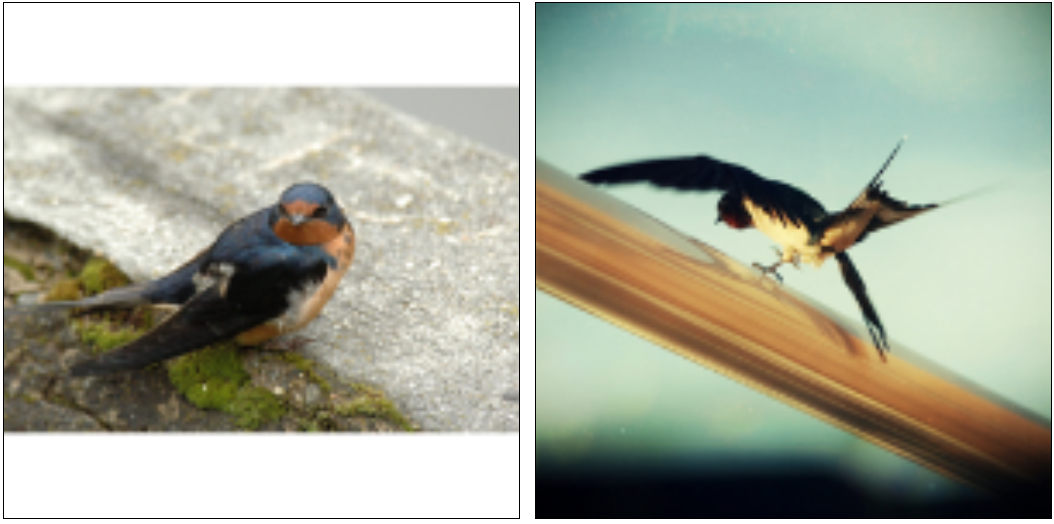}

\end{tabular}
\end{tabular} 
\caption{Left-- difficult re-identification cases in the CUHK03 dataset \cite{li2014deepreid}. The pairs in the upper row show very similar images depicting different persons, the pairs in the lower row show dissimilar images depicting the same person. Right -- analogous cases for the CUB-Birds dataset for the fine-grained classification~\cite{Wah11}. There is a clear similarity between the challenges posed by the two tasks. Both re-identification and fine-grained classification deal with strong viewpoint variations, and often need to focus on small-scale fragments in order to distinguish subjects/classes. At the same time the re-identification task has a greater degree of alignment, and we therefore suggest a modification of the bilinear CNN exploiting the presenсe of such weak alignment in the re-identification case.}
\label{fig:teaser}
\end{figure*}

The task of person re-identification is drawing the ever-increasing attention from the computer vision and the visual surveillance communities. This is because of the inherent difficulty of the task paired with the fact that medium-sized training datasets have become available only recently. The task also has a clear practical value for automated surveillance systems.
Despite a long history of research on re-identification \cite{yi2014deep, ma2012bicov, DBLP:journals/cviu/BazzaniCM13, li2015cross,prosser2010person, kuo2013person,roth2014mahalanobis,hirzer2012person,paisitkriangkrai2015learning,ma2012local,liao2015person,li2014deepreid,ahmed2015improved,chen2015deep}, the accuracy of the existing systems is often insufficient for the full automation of such application scenarios, which stimulates further research activity. The main confounding factor is the notoriously high variation of the appearance of the same person (even at short time spans) due to pose variations, illumination variation, background clutter, complemented by the high number of individuals wearing similar clothes that typically occur in the same dataset.

In this work, we follow the line of work that applies deep convolutional neural networks (CNNs) and embedding learning to the person re-identification task. Our aim is an architecture that can map (embed) an image of a detected person to a high-dimensional vector (descriptor) such that a simple metric such as Euclidean or cosine distance can be applied to compare pairs of vectors and reason about the probability of two vectors to describe the same person. Here, we avoid the approach taken in several recent works \cite{ahmed2015improved} that train a separate multi-layer network to compute the distance between a pair of descriptors, since such methods do not scale well to large datasets, where the ability to perform fast search requires the use of a simple metric.

The choice of the convolutional architecture for embedding in the case of person re-identification is far from obvious. In particular, ``standard'' architectures that combine convolutional layers followed by fully-connected layers such as those used for image classification or face embedding can fail to achieve sufficient invariance to strong 3D viewpoint changes as well as to non-rigid articulations of pedestrians, given the limited amount of training data typical for re-identification tasks and datasets. 

Here, we propose a person re-identification architecture that is based on the idea of bilinear convolutional networks (bilinear CNNs) \cite{lin2015bilinear} that was originally presented for fine-grained classification tasks and later evaluated for face recognition \cite{roychowdhury2015face}. We note that the task of person re-identification shares considerable similarity with fine-grained categorization (\fig{teaser}), as the matching process in both cases often needs to resort to the analysis of fine texture details and parts that are hard to localize.  Bilinear CNNs, however, rather radically discard spatial information in the process of the bilinear pooling. While this may be justified for fine-grained classification problems such as bird classification, the variability of geometric pose and viewpoints in re-identification problems is more restricted. Overall, the multi-region bilinear CNNs can be regarded as a middle ground between the traditional CNNs and the bilinear CNNs. In the experiments, we show that such a compromise achieves an optimal performance across a range of person re-identification benchmarks, while also performing favorably compared to previous state-of-the-art. The success of our architecture confirms the promise hold by deep architectures with multiplicative interactions such as bilinear CNNs and our multi-region bilinear CNNs for hard pattern recognition tasks.

\begin{figure*}
\begin{center}

\begin{tabular}{ c c }

\cincludegraphics[width=0.4\textwidth]{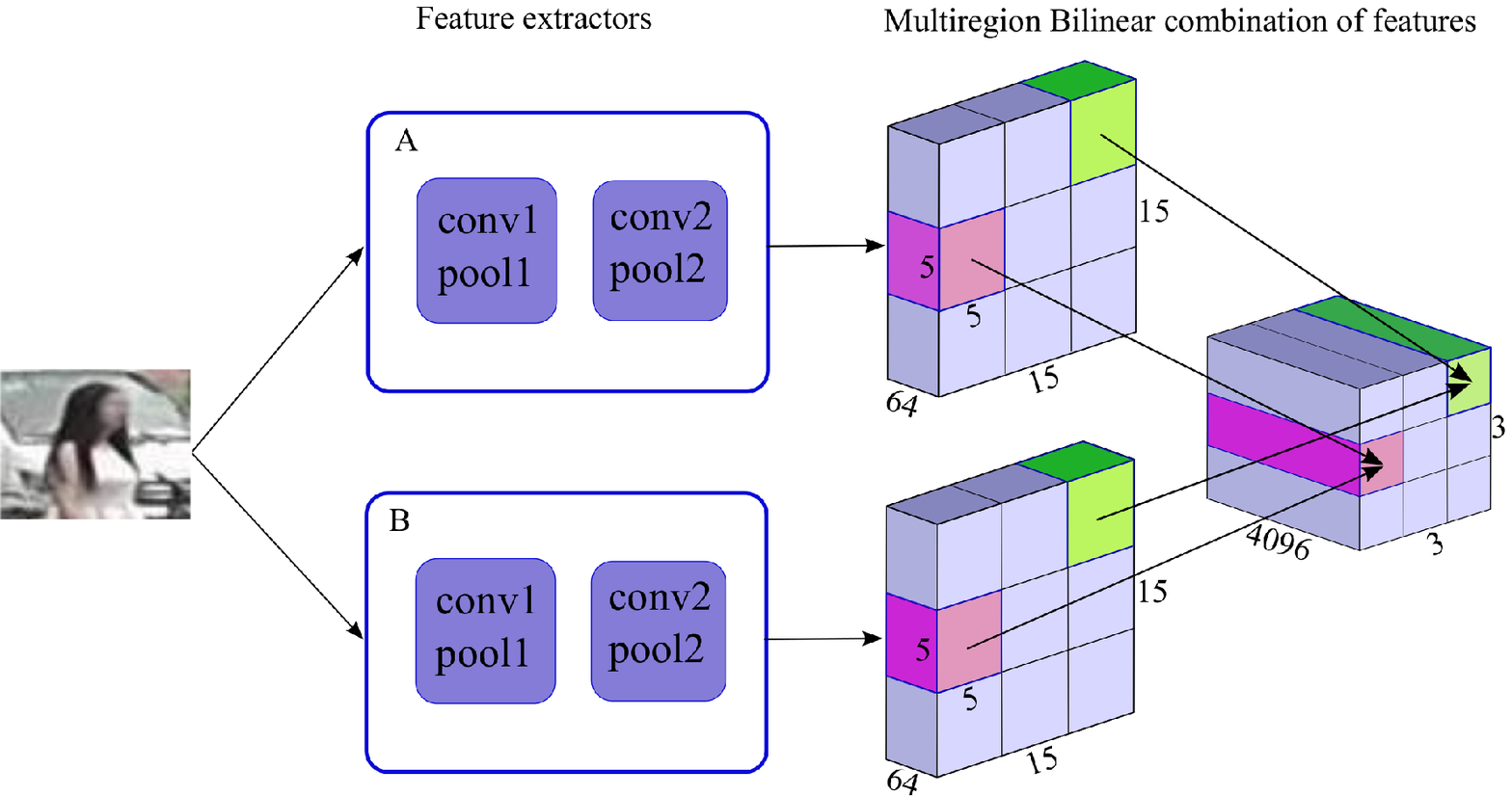}

&

\cincludegraphics[width=0.4\textwidth]{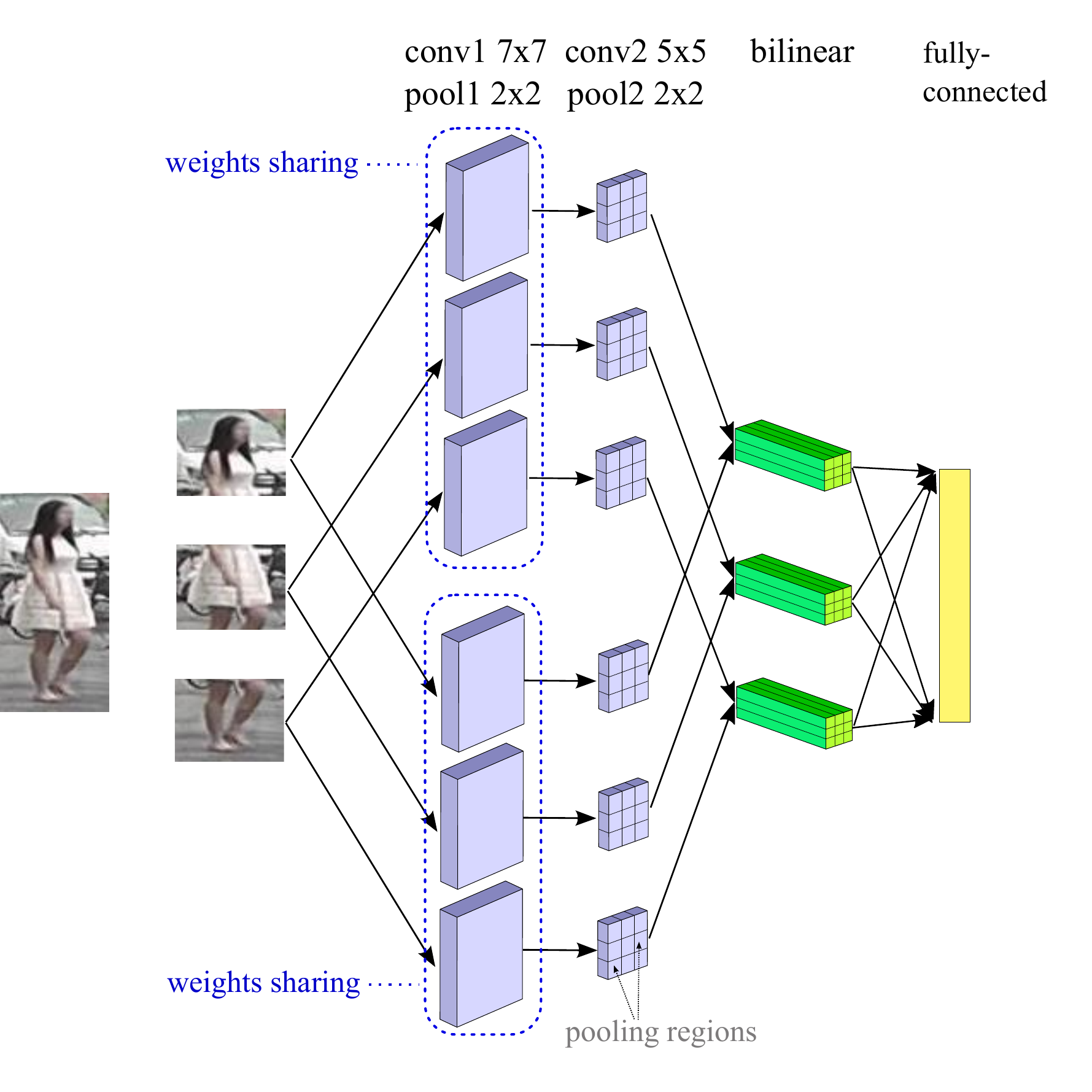}

\\
(a)&(b)
\end{tabular}
\caption{The proposed architecture for person re-identification: (a) - multi-region bilinear sub-network used for each of the three parts of the input image, (b) - the whole multi-region Bilinear CNN architecture that uses bilinear pooling over regions rather than the entire image. The new architecture achieves state-of-the-art performance over a range of benchmark datasets.}

\label{fig:architecture}

\end{center}

\end{figure*}

\section{Related work}

\textbf{Deep CNNs for Re-Identifications.} Several CNN-based methods for person re-identification have been proposed recently\cite{li2014deepreid, yi2014deep, ahmed2015improved, chen2015deep, VariorHW16, VariorSLXW16,SuZX0T16, LiuFQJY17, XiaoLOW16}. Yi~\etal~\cite{yi2014deep} were among the first to evaluate ``siamese'' architectures that accomplishes embedding of pedestrian images into the descriptor space, where they can be further compared using cosine distance. In \cite{yi2014deep}, a peculiar architecture specific to pedestrian images is proposed that includes three independent sub-networks corresponding to three regions (legs, torso, head-and-shoulders). This is done in order to take into account the variability of the statistics of textures, shapes, and articulations between the three regions. Our architecture includes the network of  Yi~\etal~\cite{yi2014deep} as its part.

Apart from \cite{yi2014deep}, \cite{li2014deepreid} and \cite{ahmed2015improved} learn classification networks that can categorize a pair of images as either depicting  the same subjects or different subjects. 
The proposed deep learning approaches \cite{ahmed2015improved, yi2014deep, li2014deepreid}, while competitive, do not clearly outperform more traditional approaches based on ``hand-engineered'' features \cite{paisitkriangkrai2015learning, zhao2014person}.

Unfortunately, when searching for matches in a dataset, the methods proposed in \cite{li2014deepreid}, \cite{ahmed2015improved} and \cite{chen2015deep} need to process pairs that include the query and every image in the dataset, and hence cannot directly utilize fast retrieval methods based on Euclidean and other simple distances. Here we aim at the approach that can learn per-image descriptors and then compare them with cosine similarity measure. This justifies starting with the architecture proposed in  \cite{yi2014deep} and then modifying it by inserting new layers.

There are several new works reporting results that are better than ours \cite{LiuFQJY17, XiaoLOW16} where additional data and/or sophisticated pre-training schemes were used, whereas we train our model from scratch on each dataset (except for CUHK01, where CUHK03 was used for pre-training).

\textbf{Bilinear CNNs.} Bilinear convolutional networks (Bilinear CNNs), introduced in \cite{lin2015bilinear} achieved state-of-the-art results for a number of fine-grained recognition tasks, and have also shown potential for face verification \cite{roychowdhury2015face}. Bilinear CNNs consists of two CNNs (where the input of these two CNNs is the same image) without fully-connected layers. The outputs of these two streams are combined in a special way via bilinear pooling. In more detail, the outer product of deep features are calculated for each spatial location, resulting in the quadratic number of feature maps, to which sum pooling over all locations is then performed. The resulting orderless image descriptor is then used in subsequent processing steps. For example, in \cite{lin2015bilinear} and \cite{roychowdhury2015face} it is normalized and fed into the softmax layer for classification. An intuition given in \cite{lin2015bilinear} is that the two CNN streams combined by bilinear operation may correspond to part and texture detectors respectively. This separation may facilitate localization when significant pose variation is present without the need for any part labeling of the training images. Our approach evaluates bilinear CNNs for the person re-identification tasks and improves this architecture by suggesting its multi-region variant.

\section{The architecture}

Our solution combines the state-of-the-art method for person re-identification (Deep Metric Learning  \cite{yi2014deep} ) and the state-of-the-art fine-grained recognition method (bilinear CNN \cite{lin2015bilinear}). Modifying the bilinear CNNs by performing multi-region pooling boosts the performance of this combination significantly. Below, we introduce the notations and discuss the components of the system in detail.


\indent\textbf{Convolutional architecture.} 
We use architecture proposed by \cite{yi2014deep} as baseline. The network incorporates independent streams, in which three overlapping parts of person images are processed separately (top, middle and bottom parts), and produces  500-dimensional descriptor as an output.

Each of the three streams incorporates two convolutional layers of size $7\times7$ and $5\times5$, followed by the rectified linear (ReLU) non-linearity and max pooling with the kernel size of two pixels and the stride of two pixels.  



\begin{figure*}
\begin{tabular}{cccc}
\includegraphics[width=0.22\textwidth]{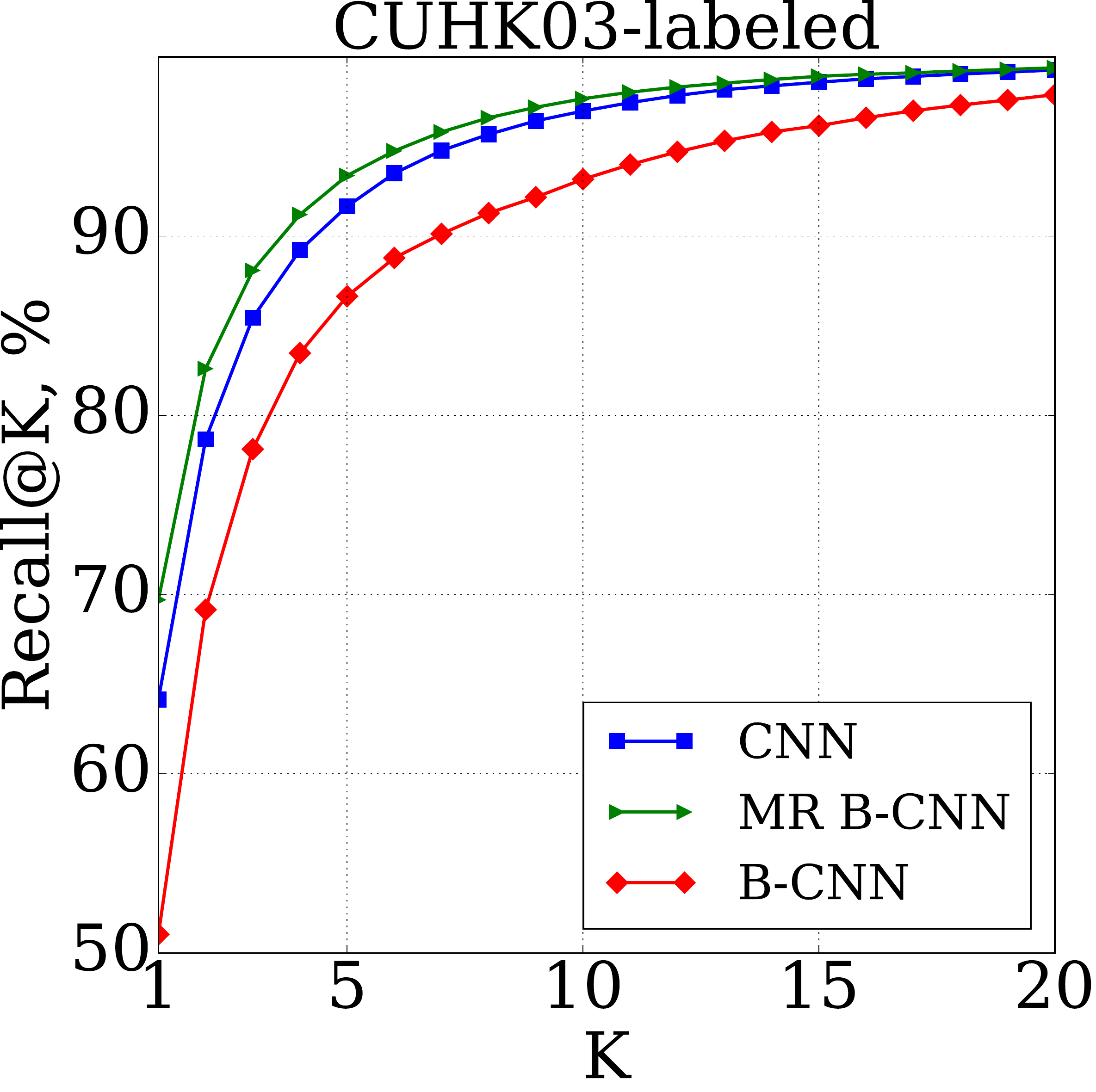}&
\includegraphics[width=0.22\textwidth]{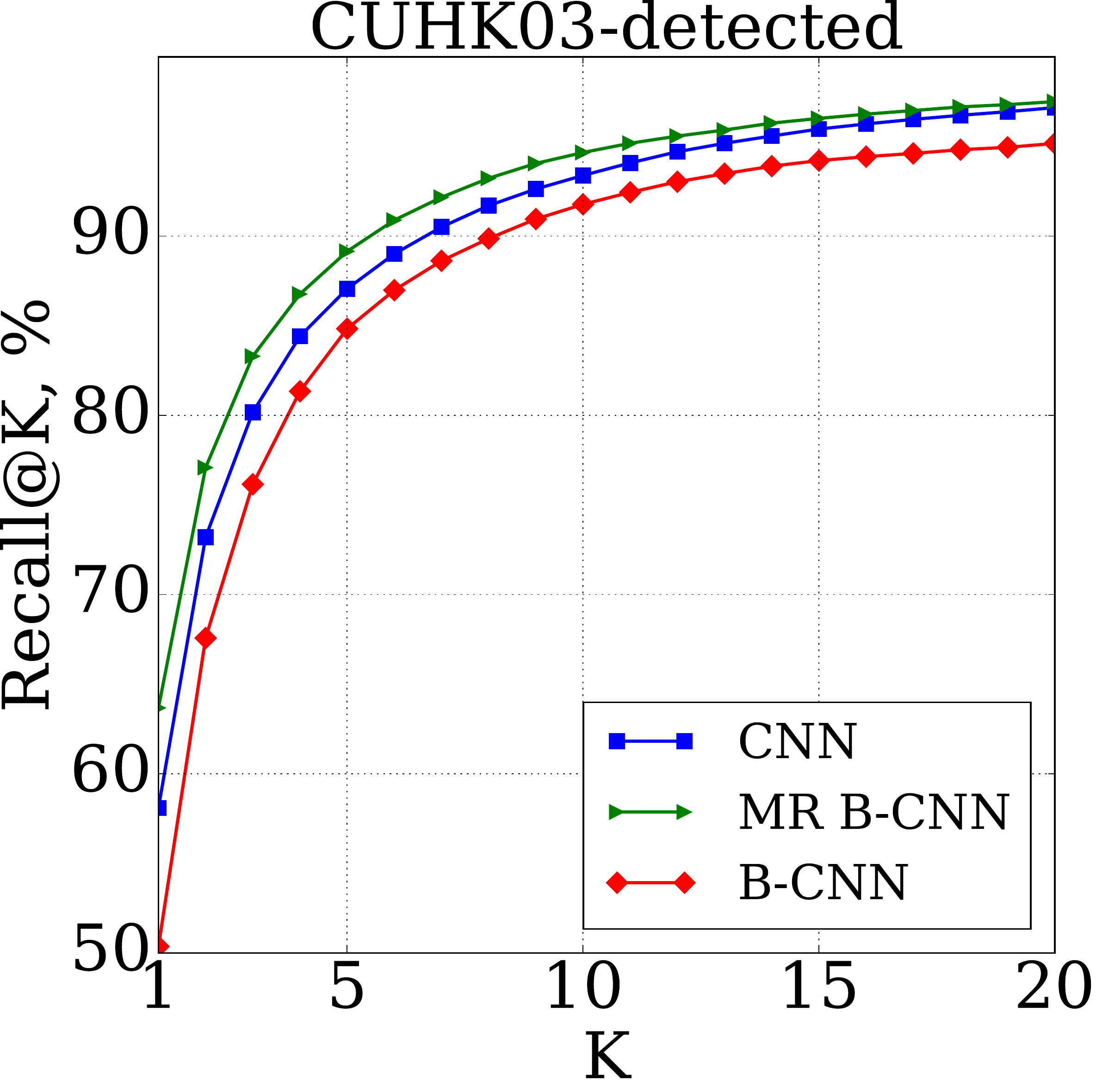}&
\includegraphics[width=0.22\textwidth]{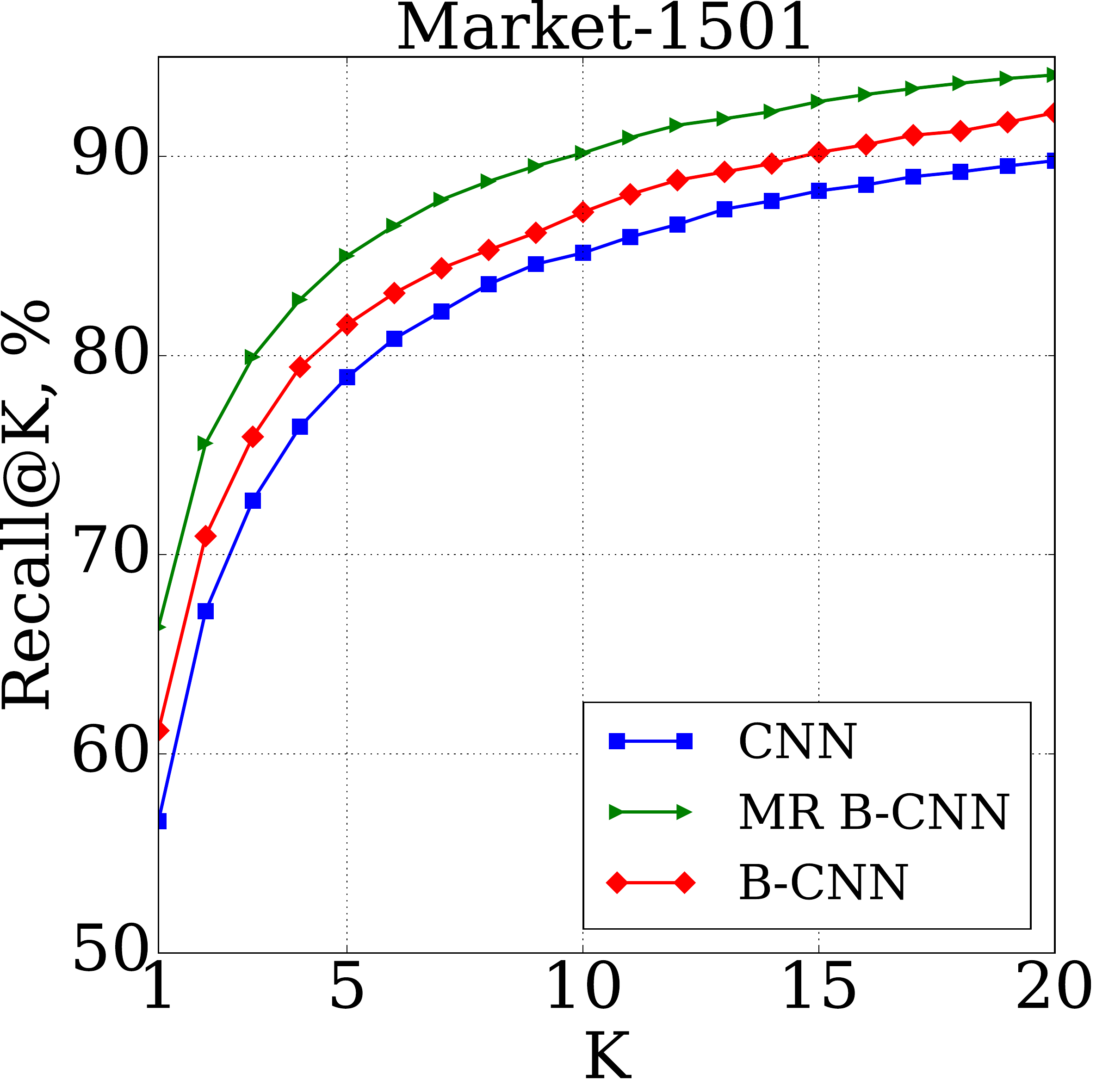}&
\includegraphics[width=0.22\textwidth]{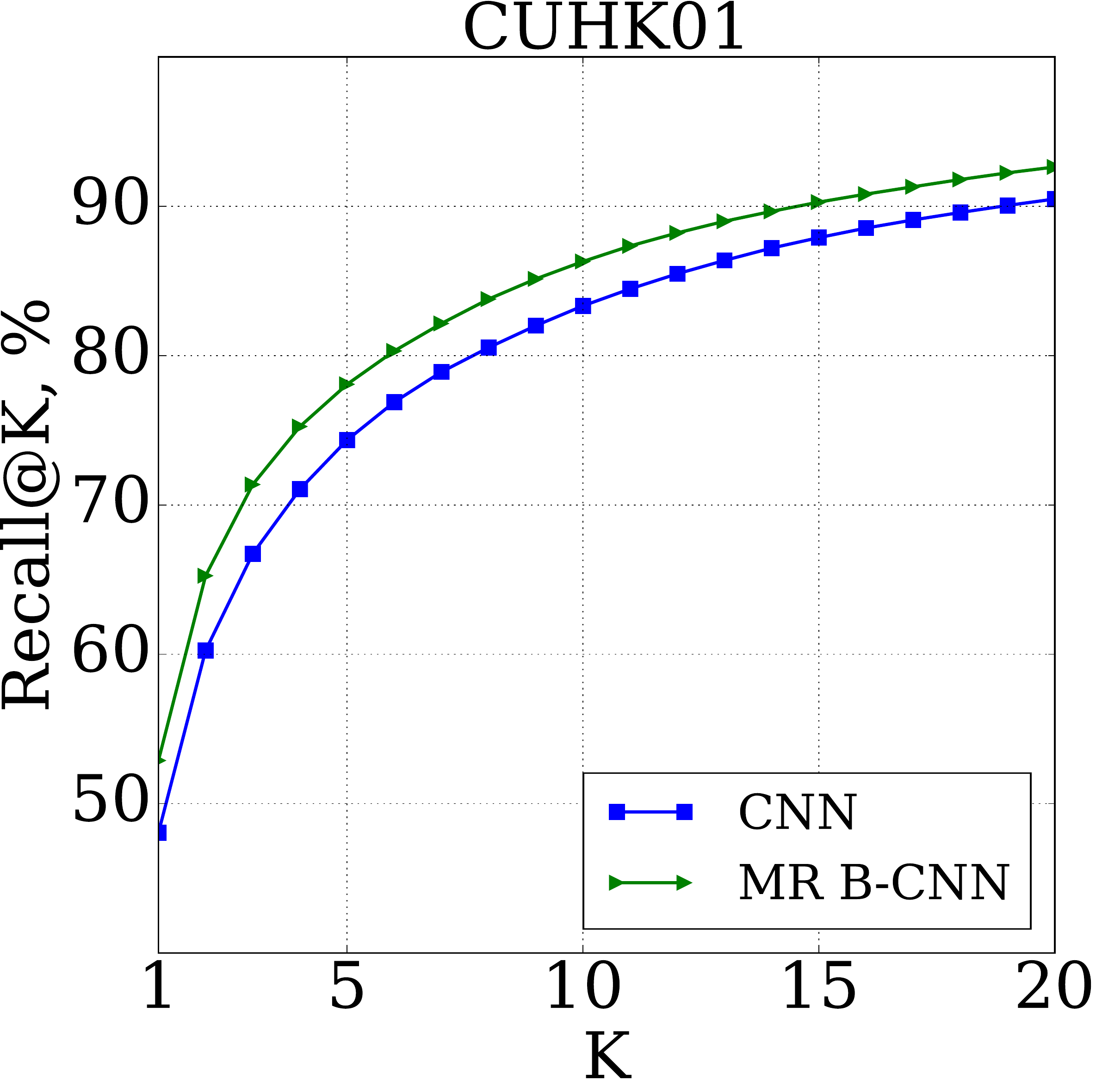}

\\
(a)&(b)&(c)&(d)
\end{tabular}
\caption{Recall@K results for the (a) CUHK03-labeled, (b) CUHK03-detected  (c) Market-1501 (d) CUHK01, datasets. MR B-CNN uniformly outperforms other architectures.}
\label{fig:recall}
\end{figure*}


\indent\textbf{Multi-region Bilinear Model.} Bilinear CNNs are motivated by the specialized pooling operation that aggregates the correlations across maps coming from different feature extractors. The aggregation however discards all spatial information that remains in the network prior to the application of the operation. This is justified when the images lack even loose alignment (as e.g.\ in the case of some fine-grained classification datasets), however is sub-optimal in our case, where relatively tight bounding boxes are either manually selected or obtained using a good person detector.  Thus some loose geometric alignment between images is always present. Therefore we modify bilinear layer and replace it with the  \textit{multi-region bilinear layer}, which allows us to retain some of the geometric information. Our modification is, of course, similar to many other approaches in computer vision, notably to the classical spatial pyramids of \cite{Lazebnik06}.
In more detail, similarly to \cite{lin2015bilinear}, we introduce the bilinear model for image similarity as follows: 
\\ $\mathcal{B} = ({f_{A}^{}}, {f_{B}^{}}, \mathcal{P}, \mathcal{S})  $, where ${f_{A}^{}}$ and ${f_{B}^{}}$ are feature extractor functions (implemented as CNNs), $\mathcal{P}$ is the pooling function, $\mathcal{S}$ is the similarity function. The feature function takes an image ${\mathcal{I}}$ at location $\mathcal{L}$ and outputs the feature of determined dimension $\mathcal{D}$ (unlike \cite{lin2015bilinear}, we use vector notation for features for simplicity): $f : \mathcal{I} \times \mathcal{L} \rightarrow \mathcal{R}_{}^{1 \times \mathcal{D}}$. In this work, two convolutional CNNs (without fully-connected layers) serve as the two feature extractors $f_{A}^{}$ and $f_{B}^{}$. For each of the two images in the pair at each spatial location, the outputs of two feature extractors $f_{A}^{}$ and $f_{B}^{}$ 
are combined using the bilinear operation \cite{lin2015bilinear}: 
\begin{equation}
 \label{eq:bilinear}
\text{bilinear}(l, im, {f_{A}^{}}, {f_{B}^{}}) = {{f_{A}^{}}(l, im)_{}^{T}{f_{B}^{}}(l, im)},
\end{equation}
where $l \in \mathcal{L}, im \in \mathcal{I}$.
Using the operation \eq{bilinear}, we compute the bilinear feature vector for each spatial location $l$ of the image $im$. If the feature extractors $f_{A}^{}$ and the $f_{B}^{}$ output local feature vectors of size $M$ and $N$ correspondingly, their bilinear combination will have size $M \times N$ or $MN \times 1$, if reshaped to the column vector. 

We then suggest to aggregate the obtained bilinear features by pooling across locations that belong to a predefined set of image regions:  ${r_{1}},  ..., {r_{R}}$, where $R$ is number of chosen regions.
After such pooling, we get the pooled feature vector for each image region $i$ (as opposed to the feature vector that is obtained in \cite{lin2015bilinear} for the whole image):
\begin{equation} \label{eq:bilinear_pooling}
\phi_{r_{i}}(im) = \phi(im_{r_{i}}) = \sum_{l \in r_{i}} \text{bilinear}(l, im, {f_{A}^{}}, {f_{B}^{}})
\end{equation}

Finally, in order to get a descriptor for image $im$, we combine all region descriptors into a matrix of size $R \times MN$:
\begin{equation} \label{eq:desc}
\phi(im) = [{\phi_{r_{1}}(im)}_{}^{T}; {\phi_{r_{2}}(im)}_{}^{T}; ... ; {\phi_{r_{R}}(im)}_{}^{T}]. 
\end{equation}


To pick the set of regions, in our experiments, we simply used the grid of equally-sized non-overlapping patches (note that the receptive fields of the units from different regions are still overlapping rather strongly). The scheme of Multi-region Bilinear CNN architecture is shown in figure~\ref{fig:architecture}a. \\
We incorporate the multi-region bilinear operation (\ref{eq:desc}) into the convolutional architecture in the following way: instead of using one sub-network for each image part, we use two feature extractors with the same convolutional architecture described above. The outputs are combined by the multi-region bilinear operation (\ref{eq:desc}) after the second convolution. Three bilinear outputs for each of the image parts are then concatenated and turned into a 500-dimensional image descriptor by an extra fully connected layer.  The overall scheme of Multi-region Bilinear CNN net for each of the two siamese sub-networks used is this work is shown in figure \ref{fig:architecture}b.
\\\indent\textbf{Learning the model.}
As in \cite{yi2014deep}, we use deep embedding learning \cite{chopra2005learning}, where multiple pedestrian images are fed into the identical neural networks and then the loss attempts to pool descriptors corresponding to the same people (\textit{matching pairs}) closer and the descriptors corresponding to different people (\textit{non-matching pairs}) apart. To learn the embeddings, we use the recently proposed Histogram loss \cite{UstinovaNIPS16} that has been shown to be effective for person re-identification task. 

\section{Experiments}



\indent\textbf{Datasets and evaluation protocols}. We investigate the performance of the CNN method and its Bilinear variant (figure \ref{fig:architecture}) for three re-identification datasets: CUHK01 \cite{LiZW12}, CUHK03 \cite{li2014deepreid} and Market-1501 \cite{zheng2015scalable}. The CUHK01 dataset contains images of 971 identities from two disjoint camera views. Each identity has two samples per camera view. We used 485 randomly chosen identities for train and the other 486 for test.  

The CUHK03 dataset includes 13,164 images of 1,360 pedestrians captured from 3 pairs of cameras. The two versions of the dataset are provided: \textit{CUHK03-labeled} and \textit{CUHK03-detected} with manually labeled bounding boxes and automatically detected ones accordingly. We provide results for both versions.

Following \cite{li2014deepreid}, we use Recall@K metric to report our results.
In more detail, the evaluation protocol accepted for the CUHK03 is the following: 1,360 identities are split into 1,160 identities for training, 100 for validation and 100 for testing. At test time single-shot Recall@K curves are calculated. Five random splits are used for both CUHK01 and CUHK03 to calculate the resulting average Recall@K. Some sample images of CUHK03 dataset are shown in figure \ref{fig:teaser}.

\begin{table}

\caption{Recall@K for the CUHK03-labeled dataset.}
\begin{tabular}{c|ccccc}
\hline
Method                & r = 1   &  r = 5   & r = 10   & r = 20 \\
\hline
FPNN \cite{li2014deepreid}  & 20.65  & 51.50    & 66.50      & 80.00   \\
LOMO+XQDA \cite{liao2015person}  & 52.20  & 82.23    & 92.14         & 96.25   \\
ImrovedDeep \cite{ahmed2015improved}& 54.74 & 86.50    & 93.88      & 98.10   \\
ME \cite{paisitkriangkrai2015learning}
                       &62.10    & 89.10    &  94.30      & 97.80  \\
DiscrNullSpace \cite{zhang2016learning}&62.55  &   90.05  & 94.80        & 98.10  \\  
\hline
 CNN                   &64.15   & 91.66    & 96.97     &99.26   \\
 MR B-CNN      & \bf{69.7}    & \bf{93.37}    &\bf{98.91}    &\bf{99.39}   \\
 \hline
\end{tabular}
\label{tab:cuhk03_labeled}
\end{table}

\begin{table}
\caption{Recall@K for the CUHK03-detected dataset. The new architecture (MR B-CNN) outperforms other methods.}
\begin{tabular}{c|ccccc}
\hline
Method                & r = 1   &  r = 5   & r = 10   & r = 20 \\
\hline
 FPNN \cite{li2014deepreid}  & 19.89  & 50.00 & 64.00         & 78.50   \\
 ImrovedDeep \cite{ahmed2015improved}& 44.96 & 76.01  & 83.47        & 93.15   \\
 LOMO+XQDA \cite{liao2015person}   & 46.25 & 78.90  & 88.55       & 94.25   \\
 DiscrNullSpace \cite{zhang2016learning}& 54.70 & 84.75   & \bf{94.80}         & 95.20 \\
 SiamLSTM \cite{VariorSLXW16}    & 57.3   &80.1    & 88.3        &-        \\
 GatedSiamCNN \cite{VariorHW16}     & 61.8    &  80.9    & 88.3         & -      \\
 \hline
 CNN                   &58.09   & 87.06    & 93.38     & 97.17  \\
 MR B-CNN        &\bf{63.67}   &\bf{89.15}    & 94.66     & \bf{97.5}   \\
 \hline
\end{tabular}
\label{tab:cuhk03_detected}
\end{table} 

We also report our results on the Market-1501 dataset, introduced in \cite{zheng2015scalable}.
This dataset contains 32,643 images of 1,501 identities, each identity is captured by from two to six cameras. The dataset is randomly divided into the test set of 750 identities and the train set of 751 identities. 


\indent\textbf{Architectures.} In the experiments, we compare the baseline CNN architecture of \cite{yi2014deep} as one of the baselines.
We also evaluate the baseline Bilinear CNN (\textit{''B-CNN''}) architecture where bilinear features are pooled over all locations for each of the three image parts. This corresponds to the formula (\ref{eq:desc}), where whole image is used for pooling.
Finally, we present the results for the Multi-region Bilinear CNN (\textit{''MR B-CNN''}) introduced in this paper (figure \ref{fig:architecture}).

\indent\textbf{Implementation details.} As in \cite{yi2014deep}, we form training pairs inside each batch consisting of 128 randomly chosen training images (from all cameras). The training set is shuffled after each epoch, so the network can see many different image pairs while training. All images are resized to height 160 and width 60 pixels. Cosine similarity is used to compute the distance between a pair of image descriptors. As discussed above, the Histogram loss \cite{UstinovaNIPS16} is used to learn the models.

\begin{table}
\caption{Recall@K for the Market-1501 dataset. The proposed architecture (MR B-CNN) outperforms other methods.}
\begin{tabular}{c|cccc}
\hline
Method                    & r = 1 & r = 5 & r = 10  &  mAP  \\
\hline
  DeepAttrDriven \cite{SuZX0T16}         & 39.4     & - & - & 19.6  \\
  DiscrNullSpace \cite{zhang2016learning}&  61.02   & - & - & 35.68 \\
  SiamLSTM \cite{VariorSLXW16}     &  61.60   & - & - & 35.31 \\
  GatedSiamCNN \cite{VariorHW16}       & 65.88  & - & - & 39.55 \\
 \hline
 CNN                  & 56.62 &  78.92 &  85.15&   32.97 \\ 
 MR B-CNN      & \bf{66.36} & \bf{85.01} & \bf{90.17}  & \bf{41.17} \\
 \hline
\end{tabular}
\label{tab:market}
\end{table} 



\begin{table}

\caption{Recall@K for the CUHK01 dataset. For CNN and MR B-CNN,  single-shot protocol with 486 queries was used. We include some results for this dataset, although we are not sure which protocol is used in \cite{zhang2016learning}. Other works use the same protocol as ours.}
\begin{tabular}{c|ccccc}
\hline
Method                & r = 1   &  r = 5   & r = 10   & r = 20 \\
\hline
ImrovedDeep \cite{ahmed2015improved} &47.53& 71.60& 80.25& 87.45\\
ME \cite{paisitkriangkrai2015learning} & 53.40 & 76.40 & 84.40 & 90.50\\
DiscrNullSpace \cite{zhang2016learning} & 69.09 &86.87 &91.77& 95.39\\
 \hline
CNN     & 48.04 &74.34 &83.33& 90.48 \\
MR B-CNN & 52.88 &78.08& 86.3& 92.63 \\
 \hline
\end{tabular}
\label{tab:cuhk01}
\end{table}

We train networks with the weight decay rate of $0.0005$. The learning rate is changing according to the ``step'' policy, the initial learning rate is set to $10^{-4}$ and it is divided by ten when the performance on the validation set stops improving (which is roughly every $100,000$ iterations). 
The dropout layer with probability of 0.5 is inserted before the fully connected layer. The best iteration is chosen using the validation set. Following \cite{ahmed2015improved}, for CUHK01 we finetune the net pretrained on CUHK03.

\indent\textbf{Variations of the Bilinear CNN architecture.} 
We have conducted a number of experiments with the varying pooling area for bilinear features (MR B-CNN), including full area pooling (B-CNN), on the CUHK03-labeled.  Here we demonstrate results for our current MR B-CNN architecture with $5\times5$ pooling area, as this architecture has been found to be the most beneficial for the CUHK03 dataset.
We also compare results for B-CNN architecture, where no spatial information is preserved. In \fig{recall}a and \fig{recall}b B-CNN is shown to be outperformed by other two architectures by a large margin. This result is not specific to a particular loss, as we observed the same in our preliminary experiments with the Binomial Deviance loss~\cite{yi2014deep}. 
The MR B-CNN architecture shows uniform improvement over baseline CNN architecture on all three datasets (\fig{recall}a,b,c,d).


\indent\textbf{Comparison with the state-of-the-art methods.} 
To our knowledge, Multi-region Bilinear CNN networks introduced in this paper outperform previously published methods on the CUHK03 (both 'detected' and 'labeled' versions), and Market-1501 datasets. Recall@K for several rank values are shown in \tab{cuhk03_labeled}, \tab{cuhk03_detected} and \tab{market} (singe query setting was used). For the Market-1501 dataset, mean average precision value is additionally shown. The results for CUHK01 are shown in \tab{cuhk01}.  

\section{Conclusion}
In this paper we demonstrated an application of new Multi-region Bilinear CNN architecture to the problem of person re-identification.
Having tried different variants of the bilinear architecture, we showed that such architectures give state-of-the-art performance on larger datasets. 
In particular, Multi-region Bilinear CNN allows to retain some spatial information and to extract more complex features, while increase the number of parameters over the baseline CNN without overfitting. We have demonstrated notable gap between the performance of the Multi-region Bilinear CNN and the performance of the standard CNN \cite{yi2014deep}. 

\FloatBarrier
{\small
\bibliographystyle{ieee}
\bibliography{refs}
}
\end{document}